\documentclass[11pt]{article}

\usepackage[preprint]{acl}
\usepackage{times}
\usepackage{latexsym}
\usepackage[T1]{fontenc}
\usepackage[utf8]{inputenc}
\usepackage{microtype}
\usepackage{inconsolata}
\usepackage{amsmath,amssymb}
\usepackage{booktabs}
\usepackage{graphicx}
\usepackage{enumitem}
\usepackage{placeins}
\usepackage{dblfloatfix}
\usepackage{float}
\usepackage{capt-of}

\title{Where Do LLM Ensemble Gains Come From?\\
Decomposing Candidate Value, Diversity, and Aggregation}

\author{
Donghwan Kim \\
Independent Researcher
}

\begin{document}
\maketitle

\begin{abstract}
LLM ensemble gains are often measured against a single baseline, but the value attributed to additional candidates can change with the comparison used.
We study an LLM aggregator under four matched conditions: \emph{Direct} (no candidates), \emph{Single-slot} (one candidate in a multi-slot interface), \emph{Duplicate} (one candidate repeated), and \emph{Multi-candidate} (separately generated candidates).
Under either non-diverse control, overall candidate gain (CIV) splits algebraically into an additional-candidate contrast (DSV) and a candidate-context contrast (CCV); the split, but not CIV, depends on the chosen control.
On MultiHopRAG model diversity ($k{=}2$), the repetition gap $\mathrm{Rep}=-8.33$\,pp $[-13.25,-3.75]$ quantifies the difference between gains measured against Duplicate and Single-slot.
On a matched FRAMES stronger-panel extension ($n{=}200$, $k{=}3$, aggregator output limit 4096), paired $\Delta\mathrm{Rep}=+9.67$\,pp $[+6.00,+13.50]$; within panels, Rep changes sign (original $-6.33$; stronger $+3.33$).
In FRAMES evidence aggregation, candidate-context contrasts remain strongly negative under both non-diverse controls.
Our results show why ensemble evaluations should state the reference control before attributing gains to additional candidates.
\end{abstract}

\section{Introduction}
\label{sec:intro}

Work on LLM ensembles usually asks whether more samples, models, or evidence raise final accuracy~\citep{wang2023selfconsistency,du2023debate,wang2024mixture,li2024more}.
Beyond majority voting, Universal Self-Consistency uses an LLM to select among free-form candidate answers~\citep{chen2023universalselfconsistency}, while LLM-Blender ranks and fuses outputs from different models~\citep{jiang2023llmblender}.
These methods show that combining candidates involves a decision about how their information is used.
Adding models is not always beneficial~\citep{li2025selfmoa,yubin2026collaboration}.
Multi-agent debate also does not consistently outperform simpler self-consistency or ensembling methods~\citep{smit2024mad}.
This makes it important to distinguish the value of additional candidates from the behavior of the system that combines them.
A separate audit finds that common diversity metrics often re-express capability, and that latent complementarity rarely becomes majority-vote gain~\citep{kim2026diversity}.
In tool-mediated industrial diagnosis, specialist information likewise need not turn into a matching gain for an LLM integrator~\citep{kim2026industrial}.

Overall accuracy alone does not reveal why an ensemble succeeds or fails.
This is a longstanding issue in ensemble learning.
Classical work relates ensemble error to diversity among predictors~\citep{krogh1994ensembles,brown2005diversity}.
More recently, \citet{choi2025debateorvote} separate the effects of voting and debate, while \citet{ali2026diversity} decomposes majority-vote gains into rescued and damaged predictions.
These studies show why aggregate performance can hide important differences between ensemble mechanisms.
We do not claim to be the first decomposition of ensemble gains.
Our contribution is a matched set of \emph{answer-level} controls for an LLM aggregator that can generate a new answer rather than only selecting a member prediction.
We test whether Duplicate and Single-slot, two plausible non-diverse controls, yield the same attribution of candidate value; they do not, and their accuracy gap changes sign between the evaluated candidate-generation configurations.

We organize the measurement around four questions:
\begin{enumerate}[leftmargin=*,itemsep=1pt,topsep=2pt]
\item Does a Multi-candidate panel help relative to \emph{Direct} solving (no candidate answers)?
\item How does Multi-candidate compare with \emph{Single-slot}, which places one candidate in the same multi-slot interface?
\item How does that comparison change when the baseline is \emph{Duplicate}, which repeats one candidate across slots?
\item What does the gap between Duplicate and Single-slot reveal about the attributed value of additional candidates?
\end{enumerate}
The central empirical finding is that \textbf{the value attributed to additional candidates depends on the non-diverse control}.

Two settings make the dependence concrete.
On MultiHopRAG model diversity ($k{=}2$), gain over Duplicate is large ($\mathrm{DSV}_\mathrm{dup}=+9.3$\,pp $[+5.0,+14.0]$) while overall candidate gain over Direct is $-3.0$\,pp (CI includes zero).
Under Single-slot, the same panel yields only $\mathrm{DSV}_\mathrm{slot}=+1.00$\,pp $[-2.33,+4.67]$, matching a negative repetition gap ($\mathrm{Rep}=-8.33$\,pp $[-13.25,-3.75]$).
On FRAMES, a matched stronger-candidate extension ($n{=}200$) raises total candidate gain by about $+20.5$\,pp while changing the estimated gain of additional candidates by $+10.3$\,pp under Single-slot but only $+0.7$\,pp under Duplicate; within panels, Rep changes sign (original $-6.33$; stronger $+3.33$), with paired $\Delta\mathrm{Rep}=+9.67$\,pp $[+6.00,+13.50]$.
Separately, evidence-mediated candidate context remains strongly negative under both controls at a 4096-token aggregator ceiling.

\paragraph{Contributions.}
\begin{enumerate}[leftmargin=*,itemsep=1pt,topsep=2pt]
\item \textbf{Controlled decomposition.}
Matched Direct / Single-slot / Duplicate / Multi-candidate conditions (experiment IDs C2 / C1$_\mathrm{slot}$ / C3b / C1) separate overall candidate gain (CIV), baseline-specific diversity contrasts ($\mathrm{DSV}_\mathrm{dup}$, $\mathrm{DSV}_\mathrm{slot}$), candidate-context contrasts (CCV), and the repetition gap ($\mathrm{Rep}$), with elementary identities $\mathrm{CIV}=\mathrm{DSV}+\mathrm{CCV}$ and $\mathrm{DSV}_\mathrm{dup}=\mathrm{DSV}_\mathrm{slot}-\mathrm{Rep}$.
The accounting is algebraic; the contribution is the empirical controls and their interpretation.
\item \textbf{Control dependence with a matched stress test.}
On MultiHop model diversity and on a matched FRAMES stronger-panel extension ($n{=}200$, 4096-token aggregation), the estimated value of additional candidates changes with the non-diverse control.
We do not claim Single-slot is uniquely neutral or isolates a pure causal diversity effect.
\item \textbf{Effects beyond baseline choice.}
Evidence $\mathrm{CCV}_\mathrm{slot}$ and $\mathrm{CCV}_\mathrm{dup}$ remain strongly negative under both controls; raw-evidence restoration and aggregator-configuration comparisons identify losses that are not explained by choosing Duplicate versus Single-slot.
\end{enumerate}

\section{Related Work}
\label{sec:related}

Self-consistency, Universal Self-Consistency, multi-agent debate, generative fusion, sampling-and-voting, and mixture-of-agents all combine multiple generations~\citep{wang2023selfconsistency,chen2023universalselfconsistency,du2023debate,jiang2023llmblender,li2024more,wang2024mixture}.
Self-MoA and related work ask whether mixing models helps when proposer quality varies~\citep{li2025selfmoa,abdulaal2025balancing}.
Multi-agent debate does not reliably beat simpler ensembling unless carefully tuned~\citep{smit2024mad}, and voting can account for much of the gain attributed to debate~\citep{choi2025debateorvote,kaesberg-etal-2025-voting}.

Classical ensemble theory relates accuracy to disagreement among predictors~\citep{krogh1994ensembles,brown2005diversity,kuncheva2003diversity}; recent LLM work likewise finds that diversity measures need not predict ensemble accuracy~\citep{kim2026diversity,ali2026diversity,chen2026cofailure}.
Kim~(2026a) examines diversity metrics as predictors of majority-vote gain, whereas we study how non-diverse aggregation controls affect the attribution of candidate value.
We do not claim the same bias-variance-covariance algebra for LLM answer accuracy; we use those results only as motivation for separating mechanism-level contrasts from a single aggregate gain.

Selector quality can matter more than diversity in generate-then-select pipelines~\citep{maryanskyy2026selection}.
Other work studies trace-level synthesis in mixture-of-agents settings~\citep{fadnavis2026beyond}.
LLM-as-a-judge is widely used~\citep{zheng2023judging}, but producing candidates and selecting among them remain distinct capabilities.
Prior work has also documented limitations of judge scores as optimization signals in a separate table-recognition setting~\citep{kim2026judge}.
Designed evidence asymmetry can help when deliberation is built for it~\citep{li2026diverseevidence}; long-context work shows that available information need not be used~\citep{liu2024lost}.
In RAG-style aggregation of multiple evidence documents, models can favor repeated paraphrases and presentation order over independent support~\citep{naphade-2026-rational}.
That line of work studies how models weight repeated evidence.
We instead measure how candidate-level value attribution changes under matched Duplicate and Single-slot \emph{answer} controls.
We add a matched Direct / Single-slot / Duplicate / Multi-candidate contrast that separates candidate-mediated context from baseline-specific diversity increments.

\section{Decomposing Ensemble Value}
\label{sec:framework}

\subsection{Four aggregation conditions}
\label{sec:four-conditions}

\begin{figure*}[t]
\centering
\includegraphics[width=0.98\linewidth]{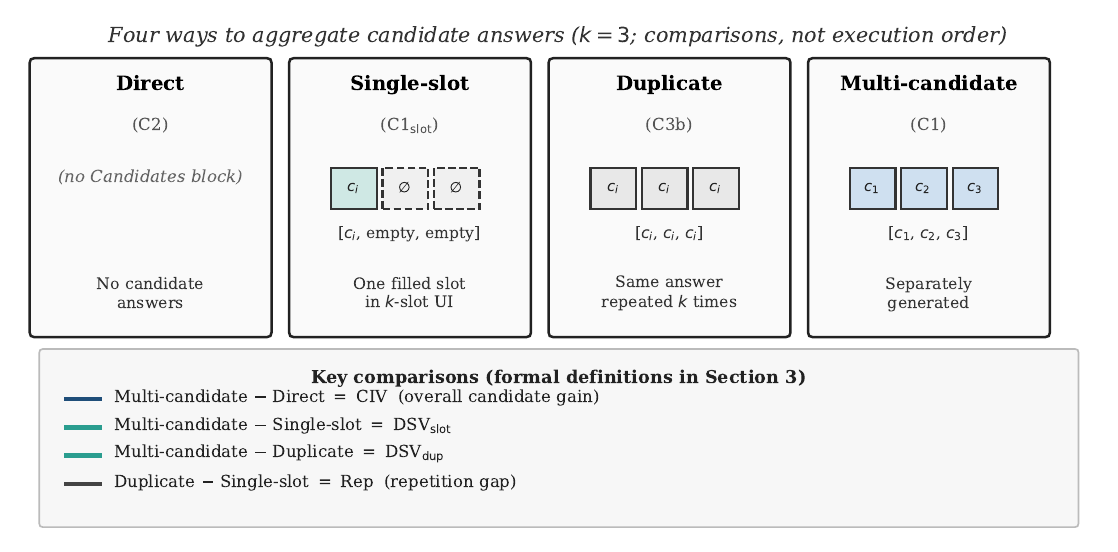}
\caption{Four aggregation conditions for panel size $k{=}3$ (comparisons, not an execution order).
Reader-facing names are primary; experiment IDs appear in parentheses.
Single-slot and Duplicate accuracies are averaged over candidate index $i$; Single-slot does not average over slot position in the primary Acc.
Separately generated Multi-candidate answers may coincide.
Neither non-diverse control identifies pure causal diversity.
Formal estimands are defined in Section~\ref{sec:framework}.}
\label{fig:decomp}
\end{figure*}

Fix one question, a panel of size $k$, and an aggregator $A$.
Figure~\ref{fig:decomp} shows four matched input constructions (reader-facing names first; experiment IDs in parentheses):
\begin{itemize}[leftmargin=*,itemsep=1pt,topsep=2pt]
\item \textbf{Direct} (C2): $A$ solves the question with no candidate answers.
In sampling/model settings the prompt is closed-book; in evidence settings the frozen primary Direct condition is a matched evidence-conditioned direct solve.
\item \textbf{Single-slot} (C1$_\mathrm{slot}$): the same $k$-slot candidate interface with one filled candidate and $k{-}1$ empty slots.
For $k{=}3$, the primary construction is $[c_i,\text{empty},\text{empty}]$ with the first slot filled and placeholder ``(No answer was provided for this slot.)''.
Reported accuracy averages over candidate index $i$, not over slot position (position and placeholder wording are later sensitivities).
\item \textbf{Duplicate} (C3b): $k$ constructed copies of one candidate, $[c_i,c_i,c_i]$ for $k{=}3$, averaged over candidate index $i$.
\item \textbf{Multi-candidate} (C1): $k$ separately generated candidate blocks.
Final answer strings may still coincide.
\end{itemize}
Aggregator prompts for Single-slot, Duplicate, and Multi-candidate are candidate-only: raw evidence is not reinjected into those arms.
Sampling, evidence, and the matched stronger-candidate extension use $k{=}3$.
The MultiHop and FRAMES model-diversity results we highlight use a pairwise ($k{=}2$) unit.

\subsection{Overall candidate gain}
\label{sec:civ}
\textbf{CIV} (candidate-information value) is Multi-candidate minus Direct:
\begin{equation}
\mathrm{CIV} = \mathrm{Acc}(\mathrm{C1}) - \mathrm{Acc}(\mathrm{C2}).
\end{equation}
It is the accuracy difference relative to the matched Direct condition.
The difference can be positive, near zero, or negative; negative CIV means the Multi-candidate setup underperforms Direct on that unit.

\subsection{Gains over alternative non-diverse controls}
\label{sec:dsv}
\textbf{DSV$_\mathrm{slot}$} is Multi-candidate minus Single-slot:
\begin{equation}
\mathrm{DSV}_\mathrm{slot} = \mathrm{Acc}(\mathrm{C1}) - \mathrm{Acc}(\mathrm{C1}_\mathrm{slot}).
\end{equation}
It asks how much accuracy changes when separately generated candidates replace one filled slot in the same interface.
Adding candidates also changes how much candidate text is present, so this is not a pure causal diversity estimand.

\textbf{DSV$_\mathrm{dup}$} is Multi-candidate minus Duplicate:
\begin{equation}
\mathrm{DSV}_\mathrm{dup} = \mathrm{Acc}(\mathrm{C1}) - \mathrm{Acc}(\mathrm{C3b}).
\end{equation}
It asks how much accuracy changes when separately generated candidates replace constructed repetitions of one answer.
The two baselines answer different questions; both are useful, and neither is uniquely ``the'' diversity effect.
In historical primary tables, bare ``DSV'' means $\mathrm{DSV}_\mathrm{dup}$.

\subsection{The repetition gap}
\label{sec:rep}
\textbf{Rep} is Duplicate minus Single-slot:
\begin{equation}
\mathrm{Rep} = \mathrm{Acc}(\mathrm{C3b}) - \mathrm{Acc}(\mathrm{C1}_\mathrm{slot}).
\end{equation}
In plain language, the repetition gap measures how much accuracy changes when one candidate is repeated instead of shown once in the multi-slot interface.
Negative Rep means Duplicate underperforms Single-slot; positive Rep means Duplicate outperforms Single-slot.
The baseline dependence of the two diversity contrasts is explicit in the identity
\begin{equation}
\mathrm{DSV}_\mathrm{slot} - \mathrm{DSV}_\mathrm{dup} = \mathrm{Rep},
\end{equation}
equivalently $\mathrm{DSV}_\mathrm{dup}=\mathrm{DSV}_\mathrm{slot}-\mathrm{Rep}$.

\subsection{Candidate-context values}
\label{sec:ccv}
After the comparisons above, we define candidate-context contrasts versus Direct:
\begin{align}
\mathrm{CCV}_\mathrm{slot} &= \mathrm{Acc}(\mathrm{C1}_\mathrm{slot}) - \mathrm{Acc}(\mathrm{C2}), \\
\mathrm{CCV}_\mathrm{dup} &= \mathrm{Acc}(\mathrm{C3b}) - \mathrm{Acc}(\mathrm{C2}).
\end{align}
Accounting identities then split the same CIV under either baseline:
\begin{equation}
\mathrm{CIV}=\mathrm{DSV}_\mathrm{dup}+\mathrm{CCV}_\mathrm{dup}=\mathrm{DSV}_\mathrm{slot}+\mathrm{CCV}_\mathrm{slot}.
\end{equation}
CIV itself does not depend on choosing Duplicate versus Single-slot; what changes is the DSV/CCV split.
We call these terms \textbf{candidate-context value} because they mix candidate quality, text presence, repetition or interface structure, and aggregator reaction.
They are accounting contrasts, not pure causal mediators.

\subsection{What the measures do not identify}
\label{sec:not-identified}
Repetition changes the input interface and can influence aggregator behavior.
Single-slot changes candidate amount and coverage relative to Multi-candidate.
Separately generated candidates can differ in quality, correctness, and content.
Neither DSV contrast alone identifies pure causal diversity.

\section{Experimental Setup}
\label{sec:setup}

\paragraph{Benchmarks and units.}
FRAMES~\citep{krishna2025frames} (sampling $n{=}200$; evidence $n{=}180$) and MultiHopRAG~\citep{tang2024multihoprag} (replication $n{=}100$).
FRAMES model diversity uses a pairwise analysis covering $30$ distinct items, not the generic FRAMES $n{=}200$ pool.
MultiHop model diversity averages three fixed $k{=}2$ model-pair comparisons within each of $n{=}100$ items.

\paragraph{Sources and aggregator.}
Sampling (S), model (M), and evidence (E) diversify candidates.
Sampling diversity uses three samples per generator, averaged across three generators.
For evidence diversity on both benchmarks, candidates come from a fixed Qwen3.8-Flash generator, and the Direct reference is the matched evidence-conditioned Direct, rather than union-evidence or closed-book Direct.
The main aggregator (``Luna'') sees parsed final answers.
Many confirmatory runs use a 1024-token output limit; later checks and the matched stronger-candidate extension use 4096.
A Qwen3-32B extension keeps candidate content fixed and swaps the aggregator \emph{configuration} (model plus decoding and reasoning settings), not a model-only intervention with equal decoding.
Verified model labels, decoding settings, and prompt locations appear in Appendix~\ref{app:repro}.
Table~\ref{tab:design} (Appendix~\ref{app:design}) summarizes datasets, panel sizes, token ceilings, and analysis status.

\paragraph{Scoring and analysis status.}
We use exact/normalized string match, then a frozen semantic judge~\citep{zheng2023judging}.
Unless noted, we report paired complete-case estimates with bootstrap 95\% CIs ($B{=}2000$, seed 2026 where specified); no $p$-values.
Historical primary analyses, later controlled extensions, and exploratory diagnostics are distinguished in Appendix~\ref{app:status} (Table~\ref{tab:status}); model settings and prompt sources appear in Appendix~\ref{app:repro}.
The historical primary S/M/E analyses use $\mathrm{DSV}=\mathrm{DSV}_\mathrm{dup}$.
We do not relabel all later results as confirmatory.

\paragraph{Human audit.}
We audited a late-stage stratified sample of $n{=}60$ judged responses (seed 2026; twelve cases from each of five strata), not a simple random sample of the completed judged universe.
Two independent raters adjudicated eight disagreements.
Versus final human labels, the judge agreed on $50/60$ ($83.3\%$; $\kappa{=}0.645$).
False rejections on this packet are mostly paraphrase or alias mismatches on FRAMES aggregator answers.
The audit does not support a CIV/DSV/CCV correction (Appendix~\ref{app:human-audit}; Tables~\ref{tab:human-audit} and~\ref{tab:human-judge-cm}).

\begin{table*}[!t]
\centering
\caption{Main Luna breakdown (pp, 95\% CIs).
DSV is gain over Duplicate ($\mathrm{DSV}_\mathrm{dup}$; C3b control).
FRAMES M uses a pairwise $k{=}2$ unit (30 distinct items), not FRAMES $n{=}200$.
Evidence CCV uses mean-duplicate C3b.}
\label{tab:main}
\small
\setlength{\tabcolsep}{5pt}
\begin{tabular}{@{}llcccc@{}}
\toprule
Bench. & Src. & CIV & DSV$_\mathrm{dup}$ & CCV$_E$ \\
\midrule
FRAMES & S & $-0.8$ [$-5.2$, $4.0$] & $0.0$ [$-2.7$, $3.0$] & n/a \\
FRAMES & M & $+0.7$ [$-10.3$, $12.0$] & $+1.8$ [$-2.5$, $6.8$] & n/a \\
FRAMES & E & $-15.9$ [$-22.6$, $-9.4$] & $+1.9$ [$-0.9$, $4.6$] & $-17.8$ [$-24.1$, $-11.3$] \\
\midrule
MultiHop & S & $-6.0$ [$-11.3$, $-0.7$] & $+4.3$ [$0.7$, $8.0$] & n/a \\
MultiHop & M & $-3.0$ [$-9.0$, $3.0$] & $+9.3$ [$5.0$, $14.0$] & n/a \\
MultiHop & E & $-13.3$ [$-20.7$, $-6.3$] & $+0.1$ [$-2.0$, $2.1$] & $-13.4$ [$-20.3$, $-6.6$] \\
\bottomrule
\end{tabular}
\end{table*}

\section{Control-Dependent Ensemble Value}
\label{sec:diverge}

Table~\ref{tab:main} reports the historical Duplicate-relative decomposition (plotted in Appendix~\ref{app:fig-main}), motivating the matched Single-slot and repetition-gap analyses that follow.
Evidence CIV (Multi-candidate minus Direct) is strongly negative on both benchmarks, while evidence $\mathrm{DSV}_\mathrm{dup}$ is near zero: most of the loss is candidate-context value, not a baseline-specific diversity increment.
This differs from settings where designed evidence asymmetry is intended to help deliberation~\citep{li2026diverseevidence}.
On MultiHop models, $\mathrm{DSV}_\mathrm{dup}=+9.3$\,pp while CIV is $-3.0$\,pp: a positive gain over Duplicate does not imply a positive candidate gain over Direct.
Sampling effects are smaller and task-dependent (near zero CIV on FRAMES; negative CIV on MultiHop).
We therefore first test whether the Duplicate baseline itself can depress accuracy, then test whether the pattern changes under a matched stronger-candidate extension.

\begin{figure*}[!t]
\centering
\includegraphics[width=0.98\linewidth]{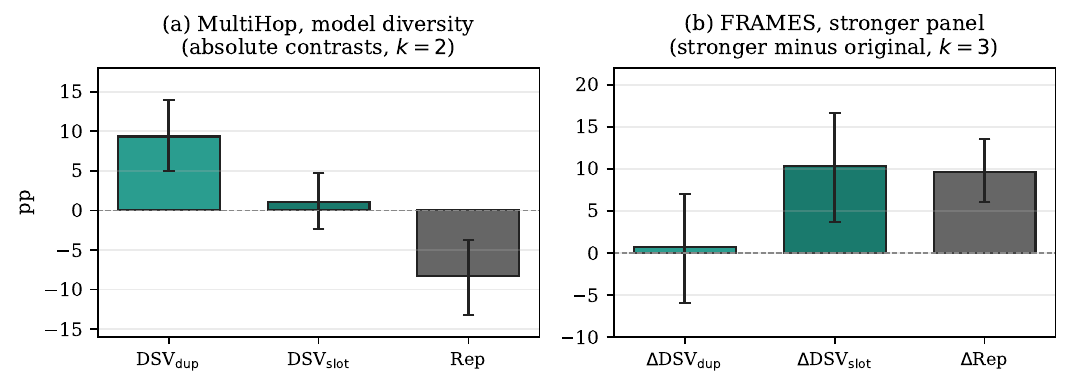}
\caption{Left: absolute MultiHop model-diversity contrasts (gain over Duplicate, gain over Single-slot, and Rep).
Right: matched FRAMES stronger-minus-original changes ($n{=}200$, 4096-token aggregation).
Both panels show paired 95\% CIs.}
\label{fig:control}
\end{figure*}

\paragraph{Control dependence.}
Under the same MultiHop model panel ($n{=}100$), gain over Single-slot is only $\mathrm{DSV}_\mathrm{slot}=+1.00$\,pp $[-2.33,+4.67]$, while the repetition gap is $\mathrm{Rep}=-8.33$\,pp $[-13.25,-3.75]$, so $\mathrm{DSV}_\mathrm{dup}=\mathrm{DSV}_\mathrm{slot}-\mathrm{Rep}$ recovers $+9.33$\,pp (Figure~\ref{fig:control}a).
Under negative Rep, $\mathrm{DSV}_\mathrm{dup}$ exceeds $\mathrm{DSV}_\mathrm{slot}$ by $-\mathrm{Rep}$.
This does not make $\mathrm{DSV}_\mathrm{dup}$ wrong; it measures a different contrast.
A CI that includes zero means the contrast is unresolved at this precision, not that the effect is zero.
The MultiHop model-diversity result is control-sensitive: $\mathrm{DSV}_\mathrm{dup}$ and $\mathrm{DSV}_\mathrm{slot}$ tell different stories.
We next ask whether the same control dependence appears when the candidate-generation configuration is strengthened on FRAMES.

\paragraph{Stronger candidates under two controls.}
Is a small Duplicate-relative DSV only a symptom of weak candidates?
A low-quality candidate panel may yield small estimates of both CIV and diversity-related contrasts, motivating a matched comparison with stronger candidates.
We therefore ask whether raising candidate quality changes the picture when the aggregator and item set stay matched.
We compare an original candidate panel with a higher-performing panel generated using different models and generation settings.
We hold the questions, panel size ($k{=}3$), aggregator (Luna), and aggregator output limit (4096 tokens) fixed, on $n{=}200$ FRAMES items (the original $95$-item cohort plus a completed nonoverlapping $105$-item complement).
The original panel uses Solar Mini4, Qwen3.8-Flash, and Haiku~4.5 with temperature $0.0$ and a $1024$-token candidate-output limit (no explicit reasoning effort set).
The stronger panel uses Gemini~3.8 Flash, Grok~4.7, and Sonnet~5.5 with temperature $0.0$, reasoning effort high, and a $65536$-token candidate-output limit.
The 4096-token aggregation limit is distinct from these candidate-generation limits.
The resulting contrast measures the value of replacing the candidate-generation configuration, not the isolated effect of model identity under equal candidate-generation compute.

\begin{figure*}[!t]
\centering
\includegraphics[width=0.98\linewidth]{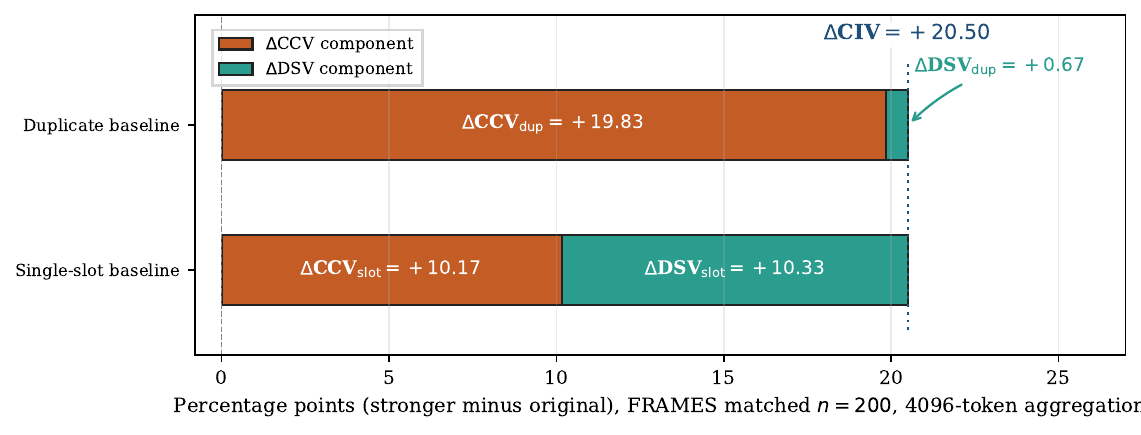}
\caption{Accounting split of the same $\Delta$CIV under Duplicate and Single-slot (not a causal path diagram).
Duplicate: $\Delta$CCV$_\mathrm{dup}=+19.83$, $\Delta$DSV$_\mathrm{dup}=+0.67$; Single-slot: $\Delta$CCV$_\mathrm{slot}=+10.17$, $\Delta$DSV$_\mathrm{slot}=+10.33$.
Both sum to $\Delta$CIV$=+20.50$; $\Delta$Rep$=+9.67$ links the DSV values. Paired CIs are in Table~\ref{tab:stronger}.}
\label{fig:alloc}
\end{figure*}

Contrasts are stronger minus original (paired bootstrap, $B{=}2000$, seed 2026).
Improving the candidate-generation configuration raises CIV by $+20.50$\,pp $[+14.00,+27.00]$.
Table~\ref{tab:abs-arms} reports the absolute arm accuracies, which clarify the within-panel signs.
Table~\ref{tab:stronger} reports the matched changes.
On the matched FRAMES items, the original Multi-candidate panel is below Direct ($41.5\%$ versus $46.5\%$; CIV $\approx-5.0$\,pp), whereas the stronger panel is above Direct ($62.0\%$ versus $46.5\%$; CIV $\approx+15.5$\,pp).
This reversal corresponds to the paired $\Delta\mathrm{CIV}=+20.50$\,pp, while the attributed additional-candidate contrast changes by $+0.67$\,pp under Duplicate versus $+10.33$\,pp under Single-slot.
The key paired contrast is $\Delta\mathrm{Rep}=+9.67$\,pp $[+6.00,+13.50]$: it directly shows that the two diversity attributions differ, rather than relying on one $\Delta$DSV CI excluding zero and another including zero (Figure~\ref{fig:control}b; Figure~\ref{fig:alloc}).
On the original panel, $\mathrm{Rep}=-6.33$\,pp $[-9.33,-3.67]$ (Duplicate worse than Single-slot).
On the stronger panel, $\mathrm{Rep}=+3.33$\,pp $[+0.83,+5.83]$ (Duplicate better than Single-slot).
Thus $\Delta\mathrm{DSV}_\mathrm{slot}=+10.33$\,pp and $\Delta\mathrm{DSV}_\mathrm{dup}=+0.67$\,pp are different splits of the same $\Delta$CIV, linked by $\Delta\mathrm{Rep}$.
Context shifts are $\Delta\mathrm{CCV}_\mathrm{dup}=+19.83$\,pp $[+15.17,+24.67]$ and $\Delta\mathrm{CCV}_\mathrm{slot}=+10.17$\,pp $[+6.33,+14.17]$.

\paragraph{Answer-count sensitivity.}
A separate controlled intervention on eligible mixed-answer cases finds that changing answer multiplicity changes aggregation accuracy by $+8.12$\,pp $[+4.99,+11.42]$ ($n{=}197$).
This supports sensitivity to answer counts but does not establish the mechanism behind the repetition-gap differences observed across candidate-generation configurations.

The $n{=}95$/$n{=}105$ rows are same-protocol diagnostics, not independent benchmarks (Appendix~\ref{app:matched}, Table~\ref{tab:cohorts}).
The MultiHop model-diversity contrast used pairwise $k{=}2$; the matched FRAMES extension uses $k{=}3$, so they are not independent replications of one intervention.
Earlier standalone-candidate evaluations used a different stack and are reported separately in Appendix~\ref{app:stronger-acc}; their estimates are not pooled with the matched comparison.
An exploratory candidate-conditioned breakdown suggests that repetition effects vary with candidate correctness; all four correctness-by-panel cells are reported in Appendix~\ref{app:rep-mech}. Correctness is post-generation and is not a causal mediator.
Slot-prompt position and placeholder variants on an $n{=}60$ subset keep positive $\Delta\mathrm{DSV}_\mathrm{slot}$ point estimates while variant-minus-primary CIs include zero (Appendix~\ref{app:slot-sens}).
The Single-slot contrast measures incremental value over a single-filled-slot control, not pure causal diversity.

\begin{table}[H]
\centering
\caption{Absolute arm accuracies (\%) and Rep (pp) on matched FRAMES $n{=}200$ (aggregator output limit 4096).
Direct (C2) is shared across panels.
Columns: Direct / Single-slot / Duplicate / Multi-candidate (experiment IDs C2 / C1$_\mathrm{slot}$ / C3b / C1).
Rep $=$ Duplicate $-$ Single-slot.
Absolute accuracies are rounded to one decimal; Rep uses the unrounded matched contrast.
Rep CIs are item-paired bootstrap ($B{=}2000$, seed 2026).}
\label{tab:abs-arms}
\scriptsize
\setlength{\tabcolsep}{2.5pt}
\resizebox{\linewidth}{!}{%
\begin{tabular}{@{}lccccc@{}}
\toprule
Panel & Direct & Single-slot & Duplicate & Multi-cand. & Rep \\
\midrule
Original & 46.5 & 45.2 & 38.8 & 41.5 & $-6.33$ [$-9.33$, $-3.67$] \\
Stronger & 46.5 & 55.3 & 58.7 & 62.0 & $+3.33$ [$+0.83$, $+5.83$] \\
\bottomrule
\end{tabular}%
}
\end{table}

\begin{table*}[!t]
\centering
\caption{Matched FRAMES stronger-minus-original effects (pp; $k{=}3$, aggregator output limit 4096).
Pooled $n{=}200$ is primary.
$\Delta$CCV$_\mathrm{dup}=+19.83$ $[+15.17,+24.67]$ and $\Delta$CCV$_\mathrm{slot}=+10.17$ $[+6.33,+14.17]$ are stated in text and Figure~\ref{fig:alloc}.
Cohort diagnostics are in Appendix~\ref{app:matched}.}
\label{tab:stronger}
\small
\setlength{\tabcolsep}{6pt}
\begin{tabular}{@{}lcccc@{}}
\toprule
 & $\Delta$CIV & $\Delta$DSV$_\mathrm{dup}$ & $\Delta$DSV$_\mathrm{slot}$ & $\Delta$Rep \\
\midrule
\textbf{Pooled $n{=}200$} & $\mathbf{+20.50}$ & $\mathbf{+0.67}$ & $\mathbf{+10.33}$ & $\mathbf{+9.67}$ \\
 & $[14.00, 27.00]$ & $[-6.00, 7.00]$ & $[3.67, 16.67]$ & $[6.00, 13.50]$ \\
\bottomrule
\end{tabular}
\end{table*}

\section{Information Loss Beyond Baseline Choice}
\label{sec:limits}

\begin{figure*}[!t]
\centering
\includegraphics[width=0.98\linewidth]{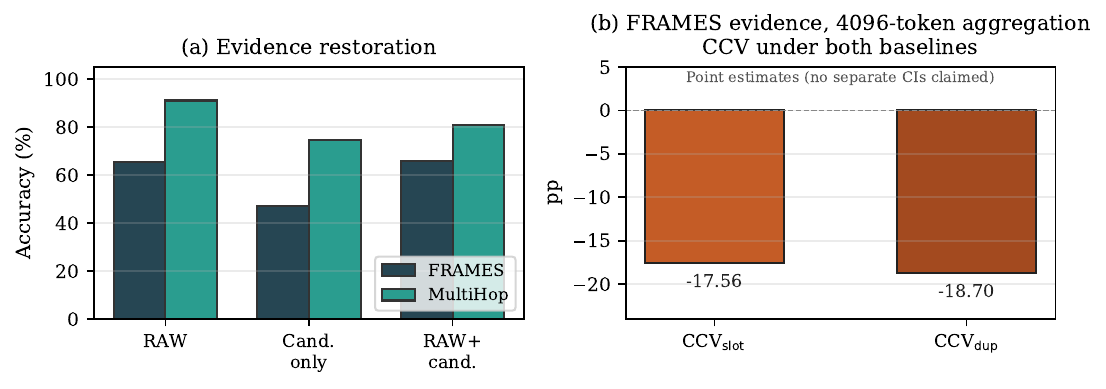}
\caption{Left: restoring raw evidence beside candidates (accuracy \%).
Right: FRAMES evidence candidate-context contrasts under Single-slot and Duplicate at 4096 tokens (point estimates).
CIV (Multi-candidate $-$ Direct) does not change between those two baselines by definition.}
\label{fig:limits}
\end{figure*}

\paragraph{Lost information through candidates.}
Making information available is not the same as extracting it into candidate answers, and extracting it is not the same as the aggregator using it~\citep{liu2024lost}.
On FRAMES, candidates-only accuracy ($47.2\%$) is far below RAW ($65.6\%$), and RAW+candidates ($65.7\%$) returns almost to RAW.
On MultiHop, recovery is only $+6.0$\,pp ($91.0$ / $74.7$ / $80.7$), so candidate context can still hurt (Figure~\ref{fig:limits}a).
We do not claim a position-bias mechanism.

\paragraph{Output-limit check.}
Re-running FRAMES evidence aggregation at 4096 tokens ($n{=}180$) lowers Multi-candidate evidence truncation from $20.4\%$ to $1.7\%$, but CIV$_E$ moves from $-15.9$\,pp to $-18.33$\,pp $[-24.26,-12.03]$.
$\mathrm{CCV}_\mathrm{slot}\approx-17.56$\,pp and $\mathrm{CCV}_\mathrm{dup}\approx-18.70$\,pp remain strongly negative under both baselines (Figure~\ref{fig:limits}b).
CIV itself is unchanged by the choice of non-diverse baseline; the empirical finding is that both candidate-context contrasts stay strongly negative.
Short outputs do not explain the evidence gap.
On robustness subsets, a second aggregator family shows the same qualitative pattern: evidence CIV stays strongly negative with near-zero evidence DSV, while model DSV can be positive (Appendix~\ref{app:kimi}, Table~\ref{tab:kimi}).

\paragraph{Control robustness.}
The remaining evidence checks distinguish information preserved in raw evidence, extracted into candidates, and used by the aggregator.
The corresponding evidence repetition gap is small at the observed point estimate, $\mathrm{Rep}\approx-1.15$\,pp $[-4.23,+2.01]$ (Appendix~\ref{app:abs-arms}).
This interval does not establish that Duplicate and Single-slot are equivalent.

\paragraph{Additional checks.}
Visible full explanations do not clearly raise $\mathrm{DSV}_\mathrm{dup}$ in our checks (FRAMES $-0.67$; MultiHop $-2.33$; both CIs include zero); a compact-rationale follow-up is similarly unresolved (Appendix~\ref{app:compact}, Table~\ref{tab:compact}).
Trace-level synthesis is methodologically different from our answer-level checks~\citep{fadnavis2026beyond}.
These secondary checks are reported in the appendix and do not change the evidence-loss result.

\newpage
\section{Discussion and Conclusion}
\label{sec:discussion}

The results imply several practices for ensemble evaluation and design.

\paragraph{Always report the baseline.}
A diversity-related gain should state whether it is measured over Single-slot or Duplicate.
On MultiHop model diversity ($k{=}2$) and the matched FRAMES stronger-panel extension ($k{=}3$), $\mathrm{DSV}_\mathrm{dup}$ and $\mathrm{DSV}_\mathrm{slot}$ tell different design stories.
If gain over Duplicate is large while gain over Single-slot is not, adding more models is not automatically justified by the Duplicate-relative number alone.

\paragraph{Do not assume repetition is neutral.}
Matched FRAMES results show that Rep can change sign between candidate-generation configurations.
This does not establish a universal causal mechanism for the sign change.

\paragraph{Candidate quality and diversity attribution are separate questions.}
Improving the candidate-generation configuration can raise CIV without producing the same increase under every diversity-related contrast.

\paragraph{Restoring raw evidence recovers most of the FRAMES gap.}
Evidence can be lost through candidate answers or aggregation; on FRAMES, restoring RAW beside candidates recovers most of the observed gap. MultiHop recovery is only partial.

\paragraph{Aggregator configuration is a distinct design variable.}
Holding candidate content fixed, the Luna-versus-Qwen3-32B evidence CIV gap is $+12.59$\,pp $[+4.07,+21.11]$ on FRAMES and $+11.0$\,pp $[+4.67,+17.67]$ on MultiHop.
These observed configuration differences show that generating candidates and realizing their information with an aggregator are distinct design questions.
They are not equal-decoding/equal-reasoning model-only interventions, do not establish Qwen Rep, and do not imply a universal ordering.

\paragraph{Cost--accuracy implications.}
Figure~\ref{fig:pareto} (Appendix~\ref{app:cost}) compares the plotted deployment cost and accuracy of the observed policies.
In its evidence-backed panel, Direct has higher plotted accuracy at lower plotted cost than the displayed M, E, and M+E candidate-based policies, illustrating why a diversity-related DSV contrast alone does not establish that extra candidate-generation and aggregation calls are worthwhile compared with Direct.
These are descriptive within-panel comparisons: policies use different token limits, and C2$_{\mathrm{PRO}}$ also changes model setup, so Figure~\ref{fig:pareto} does not establish a general cost-optimal or equal-reasoning-budget policy.

\paragraph{Conclusion.}
For LLM ensembles, evaluation should state which counterfactual defines the measured gain, how much candidate information survives the interface, and which effects persist across controls.
Reporting CIV, baseline-specific DSV/CCV, and Rep makes those distinctions visible.

\FloatBarrier
\section{Limitations}
\label{sec:limitations}

We use selected short-answer benchmarks and mainly answer-level aggregation.
Visible-explanation checks do not cover hidden traces.
Single-slot (C1$_\mathrm{slot}$) changes coverage as well as distinctness; Duplicate (C3b) changes repetition; neither is a pure causal diversity estimand.
The matched stronger-versus-original contrast changes candidate models and generation settings together, so it is not a model-only effect at equal candidate-generation compute.
Evidence-diversity findings use a single fixed candidate generator on both benchmarks, so their transfer to other generators is untested.

Headline tests primarily use one aggregator (Luna); Qwen CIV gaps do not establish Qwen Rep.
The human audit is a stratified $n{=}60$ packet and does not demonstrate absence of arm-differential judge error on the matched Single-slot/Duplicate cells; we do not numerically correct main effects.
Aggregation truncation was tracked, and the frozen inclusion policy was retained rather than switching to success-only accuracy.
We do not claim a complete candidate-generation empty/parse-failure comparison between panels beyond the disclosed generation-config differences and aggregation-arm health checks.

Candidate-conditioned Rep analyses are exploratory and condition on a post-generation variable.
Slot position/placeholder variants ($n{=}60$) do not prove equivalence of prompt forms.
Later extensions were motivated by earlier findings; they are not all confirmatory.
Public OpenRouter slugs and decoding settings appear in Appendix~\ref{app:repro}; vendor-internal build IDs remain unresolved.

\bibliography{references}

\clearpage
\appendix

\section{Historical Primary Split Figure}
\label{app:fig-main}
\begin{figure*}[!htbp]
\centering
\includegraphics[width=\textwidth]{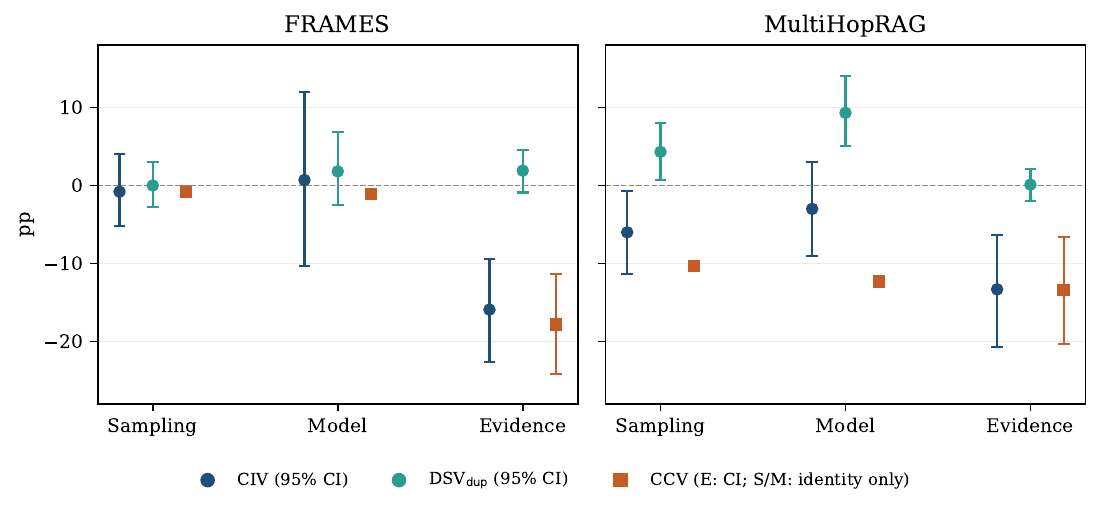}
\caption{\textbf{Appendix A. Historical primary split figure.}
DSV $=\mathrm{DSV}_\mathrm{dup}$.
CIV and DSV$_\mathrm{dup}$ show paired 95\% CIs.
Evidence CCV has a CI; sampling/model CCV markers are identity points without separate CIs.}
\label{fig:main}
\end{figure*}

Figure~\ref{fig:main} visualizes the historical primary S/M/E split summarized in Table~\ref{tab:main}.
It provides a cross-source view of CIV, DSV$_\mathrm{dup}$, and evidence CCV; Table~\ref{tab:main} remains the source of the reported estimates.

\section{Experiment Design Summary}
\label{app:design}
\begin{center}
\footnotesize
\setlength{\tabcolsep}{1.5pt}
\begin{tabular}{@{}llccccc@{}}
\toprule
Dataset & Source & $n$ & $k$ & Tokens & Unit & Status \\
\midrule
FRAMES & S & 200 & 3 & 1024 & item & Primary (A) \\
FRAMES & M & 30 & 2 & 1024 & pairwise & Primary (A) \\
FRAMES & E & 180 & 3 & 1024 & item & Primary (A) \\
MultiHop & S/M/E & 100 & 2/3 & 1024 & item/pair & Primary (A) \\
FRAMES & stronger & 200 & 3 & 4096 & item & Extension (B) \\
\bottomrule
\end{tabular}
\captionof{table}{Core experiment design (verified against frozen configs).
FRAMES M uses 30 distinct items with pairwise comparison/averaging, not the generic FRAMES $n{=}200$ pool.
The matched stronger-candidate row combines the original $95$-item cohort and a nonoverlapping $105$-item complement; it is a later controlled extension, not two independent benchmarks.}
\label{tab:design}
\end{center}
For FRAMES evidence, the frozen 180-item panel assigns three per-item budget-matched Wikipedia evidence packets using source-disjoint allocation where feasible; each candidate is generated from one packet, whereas matched Direct receives their deterministic interleaving.

\section{Analysis-Status Transparency}
\label{app:status}
\begin{center}
\scriptsize
\setlength{\tabcolsep}{3pt}
\begin{tabular}{@{}p{0.47\linewidth}p{0.215\linewidth}p{0.215\linewidth}@{}}
\toprule
Analysis & Status & Notes \\
\midrule
S/M/E: CIV, DSV$_\mathrm{dup}$, CCV & Confirmatory (A) & Historical primary \\
4096 checks; RAW+candidates; stronger-panel extension; MultiHop Rep/slot; controlled answer-multiplicity; Qwen; explanations & Extension (B) & Fixed before outcomes \\
Slot $n{=}60$; compact rationale & Diagnostic (C) & Not primary \\
Human audit $n{=}60$ & Completed (B) & Stratified; no recalibration \\
\bottomrule
\end{tabular}
\captionof{table}{Analysis hierarchy.}
\label{tab:status}
\end{center}

\section{Matched Cohort Diagnostics}
\label{app:matched}
\begin{center}
\scriptsize
\setlength{\tabcolsep}{2.5pt}
\resizebox{\linewidth}{!}{%
\begin{tabular}{@{}lcccc@{}}
\toprule
Cohort & $\Delta$CIV & $\Delta$DSV$_\mathrm{dup}$ & $\Delta$DSV$_\mathrm{slot}$ & $\Delta$Rep \\
\midrule
Original cohort $n{=}95$ & $+17.89$ & $-1.05$ & $+10.53$ & $+11.58$ \\
 & $[9.47, 26.32]$ & $[-10.18, 8.42]$ & $[1.75, 20.00]$ & $[6.67, 16.84]$ \\
Compl.\ $n{=}105$ & $+22.86$ & $+2.22$ & $+10.16$ & $+7.94$ \\
 & $[13.33, 31.43]$ & $[-6.67, 10.48]$ & $[0.95, 18.73]$ & $[2.86, 13.33]$ \\
\bottomrule
\end{tabular}%
}
\captionof{table}{Matched FRAMES stronger-minus-original cohort diagnostics (pp).
Same protocol cohorts, not independent benchmarks.}
\label{tab:cohorts}
\end{center}
The complement completion recorded 13 truncations; scoring retains the matched denominator.

\section{Human Audit Details}
\label{app:human-audit}
Late-stage stratified sample: twelve cases from each of five judged strata (standard S/M/E, factorial evidence-layout manipulation, MultiHop replication, Kimi robustness, and alternate Direct configuration C2-Pro; seed 2026), drawn from the then-available judged universe and not redrawn after later call completion.
Two raters labeled Correct / Incorrect / Ambiguous, blinded to judge decisions; eight disagreements were adjudicated.
\begin{center}
\small
\setlength{\tabcolsep}{4pt}
\begin{tabular}{@{}p{0.42\linewidth}p{0.48\linewidth}@{}}
\toprule
Quantity & Value \\
\midrule
Rater agreement (3-class) & $52/60{=}86.7\%$; $\kappa{=}0.746$ \\
Human vs.\ judge agreement & $50/60{=}83.3\%$; $\kappa{=}0.645$ \\
TP / FP / TN / FN & $16$ / $0$ / $34$ / $10$ \\
\bottomrule
\end{tabular}
\captionof{table}{Human audit summary (stratified $n{=}60$).}
\label{tab:human-audit}
\end{center}
\begin{center}
\small
\begin{tabular}{@{}lcc@{}}
\toprule
 & Human Correct & Human Incorrect \\
\midrule
Judge Correct & TP${=}16$ & FP${=}0$ \\
Judge Incorrect & FN${=}10$ & TN${=}34$ \\
\bottomrule
\end{tabular}
\captionof{table}{Confusion matrix (positive $=$ judge Correct; reference $=$ final human).}
\label{tab:human-judge-cm}
\end{center}
All ten FN cases on this packet are FRAMES aggregator paraphrase, alias, or unit-equivalent mismatches.
One FRAMES audit case (A05) was flagged for external AI assistance during labeling; excluding it gives human--judge agreement $49/59$ (vs.\ $50/60$ overall), with specificity unchanged.
The audit findings are reported without adjusting the primary effect estimates.

\section{Cost Summary}
\label{app:cost}
\begin{figure*}[!htbp]
\centering
\includegraphics[width=\textwidth]{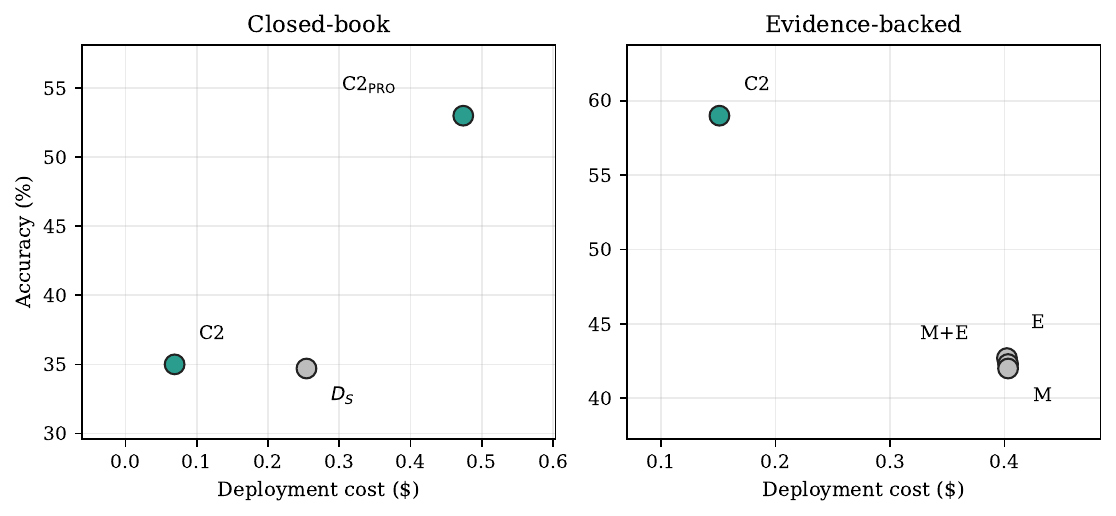}
\caption{\textbf{Appendix F. Cost summary.}
Accuracy versus deployment dollar cost.
Costs are total deployment costs (USD) on a 100-item FRAMES overlap subset of the historical panels; judge costs are excluded.
Because policy construction and model averaging differ from Table~\ref{tab:main}, these accuracies are not directly comparable to Table~\ref{tab:main} estimates.
Here $D_S$ denotes sampling diversity: three samples per generator, averaged across the three generators.
Filled markers are non-dominated within each panel.
Policies use different token limits; C2$_{\mathrm{PRO}}$ also changes model setup.}
\label{fig:pareto}
\end{figure*}

Figure~\ref{fig:pareto} presents the cost--accuracy view for the evaluated policies.
It summarizes the historical primary policy evaluations, not the matched stronger-candidate extension.
It is descriptive because policies use different token limits and C2$_{\mathrm{PRO}}$ also changes the model setup.

\section{Kimi Robustness (Selective)}
\label{app:kimi}
\begin{center}
\small
\begin{tabular}{@{}llrr@{}}
\toprule
Bench. & Estimand & Est.\ (pp) & 95\% CI \\
\midrule
FRAMES & CIV$_E$ & $-15.7$ & [$-28.7$, $-2.8$] \\
FRAMES & DSV$_E$ & $+1.5$ & [$-3.7$, $+7.4$] \\
FRAMES & DSV$_M$ & $+6.7$ & [$0.0$, $+13.3$] \\
MultiHop & CIV$_E$ & $-15.0$ & [$-27.2$, $-3.9$] \\
MultiHop & DSV$_E$ & $+0.2$ & [$-2.4$, $+2.6$] \\
MultiHop & DSV$_M$ & $+5.6$ & [$+0.6$, $+10.6$] \\
\bottomrule
\end{tabular}
\captionof{table}{Kimi robustness subsets (not the main Luna confirmatory panel).}
\label{tab:kimi}
\end{center}

\section{Task-wise Stronger-Candidate Accuracies}
\label{app:stronger-acc}
\begin{center}
\scriptsize
\setlength{\tabcolsep}{2.5pt}
\resizebox{\linewidth}{!}{%
\begin{tabular}{@{}llcc@{}}
\toprule
Bench. & Candidate configuration & Acc. & vs.\ Luna Direct (C2) \\
\midrule
FRAMES & Luna Direct (C2) & 44.5\% & n/a \\
FRAMES & Gemini High & 61.5\% & higher \\
FRAMES & Grok High & 52.0\% & higher \\
FRAMES & Sonnet High & 50.0\% & higher \\
MultiHop & Luna Direct (C2) & 87.0\% & n/a \\
MultiHop & Gemini High & 95.0\% & higher \\
MultiHop & Grok High & 85.0\% & lower \\
MultiHop & Sonnet High & 72.0\% & lower \\
\bottomrule
\end{tabular}%
}
\captionof{table}{Standalone task accuracy vs.\ Luna Direct (C2) on the earlier standalone-candidate evaluation stack (FRAMES Luna Direct $=44.5\%$).
This is not the matched stronger-panel Direct (C2) in Table~\ref{tab:abs-arms} ($46.5\%$ at aggregator output limit 4096).}
\label{tab:taskacc}
\end{center}

\section{Compact Rationale (Exploratory)}
\label{app:compact}
Answer-only (A0) places only the parsed answer in each candidate slot. Compact visible rationale (R1) appends a sentence-boundary-respecting prefix of the visible explanation, truncated to $K{=}200$ tokens with \texttt{o200k\_base}; Full visible explanation (R2) appends the complete visible explanation.
This exploratory comparison reuses stored stronger-panel candidate responses and their visible explanations, without new candidate generation or model-based summarization. The original candidate panel fell below the frozen 95\% visible-explanation coverage requirement (FRAMES 29\%; MultiHopRAG 47\%) and was excluded.
Here DSV is Duplicate-relative, $\mathrm{DSV}_\mathrm{dup}=\mathrm{Acc}(\mathrm{C1})-\mathrm{mean}_i\,\mathrm{Acc}(\mathrm{C3b}_i)$, so the rows are changes in that contrast between representations; $K$ is rationale-prefix length, not panel size $k$.
\begin{center}
\scriptsize
\setlength{\tabcolsep}{2.5pt}
\resizebox{\linewidth}{!}{%
\begin{tabular}{@{}lcc@{}}
\toprule
Contrast & FRAMES & MultiHop \\
\midrule
R1$-$A0 DSV & $-1.50$ [$-8.17$, $+5.00$] & $+2.33$ [$-4.00$, $+8.67$] \\
R2$-$R1 DSV & $+0.83$ [$-5.17$, $+6.67$] & $-4.67$ [$-10.67$, $+0.67$] \\
\bottomrule
\end{tabular}%
}
\captionof{table}{Post-hoc exploratory compact-rationale contrasts ($K{=}200$).
This exploratory analysis was defined after the full-explanation results. The primary estimates and conclusions remain unchanged.}
\label{tab:compact}
\end{center}

\section{Repetition Gaps Across Settings}
\label{app:abs-arms}
Table~\ref{tab:abs-arms} reports within-panel accuracies on the matched FRAMES stack.
Across settings with matched C3b and C1$_\mathrm{slot}$, Rep varies in sign and magnitude: MultiHop model diversity $\mathrm{Rep}=-8.33$\,pp $[-13.25,-3.75]$ ($k{=}2$); FRAMES evidence $\mathrm{Rep}\approx-1.15$\,pp $[-4.23,+2.01]$ at aggregator output limit 4096 (CI includes zero).
These rows differ in task, $k$, and panel, so they illustrate sensitivity rather than a universal law.

\section{Slot-Prompt Sensitivity}
\label{app:slot-sens}
On a frozen $n{=}60$ subset of the matched FRAMES stack, primary $\Delta\mathrm{DSV}_\mathrm{slot}=+7.50$\,pp $[-2.22,+17.50]$.
Filling the last slot instead yields $+3.89$\,pp $[-6.67,+13.89]$; an alternate placeholder ``[EMPTY SLOT]'' yields $+5.00$\,pp $[-3.90,+14.44]$.
Paired differences versus primary include zero (pos2$-$primary $-3.61$ $[-11.67,+4.17]$; altph$-$primary $-2.50$ $[-10.28,+5.56]$).
Directional estimates remain positive; this does not establish equivalence, so C1$_\mathrm{slot}$ is not uniquely neutral.

\section{Exploratory Candidate-Conditioned Rep}
\label{app:rep-mech}
Using frozen candidate correctness labels on the matched FRAMES panels, define candidate-wise $r_i=\mathrm{Correct}(\mathrm{C3b}_i)-\mathrm{Correct}(\mathrm{C1}_{\mathrm{slot},i})$.
Item-clustered means (pp; $B{=}2000$, seed 2026; CI withheld if fewer than 20 items):
original correct $+2.38$ $[0.00,+7.53]$ ($73$ items / $126$ obs);
original incorrect $-8.65$ $[-13.30,-5.74]$ ($183$ / $474$);
stronger correct $+5.81$ $[+2.28,+10.84]$ ($146$ / $327$);
stronger incorrect $+0.37$ $[-2.84,+6.05]$ ($135$ / $273$).
Correct-candidate counts also shift between panels (original mass on 0 correct; stronger mass on 2--3).
These patterns are descriptive associations under a post-generation covariate, not an identified mediator.

\section{Reproducibility Notes}
\label{app:repro}
Model identifiers and decoding configurations were verified against recorded request configurations and execution artifacts. Public OpenRouter slugs are listed below; vendor-internal build IDs are not exposed by the provider.

{\small
\begin{description}[style=nextline,leftmargin=1.1em,itemsep=0pt,parsep=0pt]
\item[Primary aggregator (Luna)] \texttt{openai/gpt-5.6-luna}; OpenAI pin via OpenRouter; temperature $0.0$; max\_tokens $1024$ (primary) / $4096$ (matched extensions); seed $2026$ where specified; request \texttt{reasoning=null} (effort not explicitly set; not claimed disabled).
\item[Qwen aggregator] \texttt{qwen/qwen3-32b}; DeepInfra route; thinking ON; temperature $0.6$; top$_p$ $0.95$; top$_k$ $20$; max\_tokens $8192$ in the wire specification; candidate content fixed.
\item[Semantic judge] \texttt{qwen/qwen3-32b} as a separate role; temperature $0.0$; max\_tokens $256$; seed $2026$; YES/NO prompt; not the Qwen aggregator configuration.
\item[Original candidates] \texttt{upstage/solar-mini4}, \texttt{qwen/qwen3.8-flash}, \texttt{anthropic/claude-haiku-4.5}; temperature $0.0$; max\_tokens $1024$; sample seeds $2026$--$2028$ where applicable; no explicit reasoning-effort override.
\item[Stronger candidates] \texttt{google/gemini-3.8-flash}, \texttt{x-ai/grok-4.7}, \texttt{anthropic/claude-sonnet-5.5}; temperature $0.0$; reasoning effort high; max\_tokens $65536$; seed $2026$; frozen provider routing.
\end{description}
}

Additional verified details:
\begin{itemize}[leftmargin=*,itemsep=0pt]
\item Panel sizes: $k{=}3$ for sampling/evidence/matched stronger aggregation; pairwise $k{=}2$ for highlighted model-diversity tables.
\item Aggregation ceilings: $1024$ for many primary runs; $4096$ for the evidence ceiling check and matched stronger-panel extension (aggregation stage, not stronger candidate generation).
\item Slot placeholder (primary): ``(No answer was provided for this slot.)''; a sensitivity uses ``[EMPTY SLOT]'' and an alternate fill position.
\item All candidate-aware aggregation conditions share one fixed aggregator template; only the candidate-slot contents differ.
\item Bootstrap: paired complete-case, $B{=}2000$, seed $2026$ where specified.
\item Scoring: exact/normalized match, then frozen semantic judge; human audit stratified $n{=}60$ as above. Truncation finish-reason and token-ceiling flags are retained; the matched stronger complement recorded truncations but kept the matched denominator. Remaining unresolved: provider-default reasoning behavior when effort is unspecified and prospective provider pins never set on some original candidate calls.
\end{itemize}

\end{document}